\documentclass[lettersize,journal]{IEEEtran}
\usepackage{amsmath,amsfonts}
\usepackage{array}
\usepackage[caption=false,font=normalsize,labelfont=sf,textfont=sf]{subfig}
\usepackage{textcomp}
\usepackage{stfloats}
\usepackage{url}
\usepackage{verbatim}
\usepackage{graphicx}
\usepackage{cite}
\usepackage{xcolor}
\usepackage{booktabs}
\usepackage{multirow}
\usepackage{multicol}
\usepackage{textcomp}
\usepackage[ruled,vlined,noend]{algorithm2e}

\begin{document}
\title{Coaching a Robotic Sonographer: Learning Robotic Ultrasound with Sparse Expert's Feedback}
% \title{Learning of Robotic Ultrasound by Coaching \vspace{-0.2cm}}
% $^{1*}$
\author{Deepak Raina$^*$, Mythra V. Balakuntala$^*$, Byung Wook Kim, Juan Wachs ~\IEEEmembership{Senior Member,~IEEE}, \\Richard Voyles \IEEEmembership{Fellow,~IEEE}
% \\
% \vspace{-1cm}
\thanks{This work was supported by National Science Foundation (NSF) USA under Grant \#2140612 and Purdue University seed grants for West Lafayette-Indianapolis campuses collaboration. (Corresponding author: Deepak Raina)}
% \thanks{All the authors are affiliated with Purdue University (PU), West Lafayette, Indiana, USA (\{draina, mbalakun, kim2986,  jpwachs, rvoyles\}@purdue.edu)}
\thanks{Deepak Raina is with the Malone Center for Engineering in Healthcare, Johns Hopkins University, Baltimore, MD 21218 USA (e-mail: draina1@jh.edu)}
\thanks{Mythra V. Balakuntala is with Nikon Research Corporation of America, Belmont, CA 94002 USA (e-mail: mythra.balakuntala@nikon.com)}
% \thanks{Byung Wook Kim is with the Department of Electrical and Computer Engineering, Purdue University, West Lafayette, IN 47907 USA (e-mail: kim2986@purdue.edu)}
\thanks{Byung Wook Kim is with the Department of Computer Engineering, Purdue University, West Lafayette, IN 47907 USA (e-mail: kim2986@purdue.edu)}
\thanks{Juan Wachs is with the School of Industrial Technology, Purdue University, West Lafayette, IN 47907 USA (e-mail: jpwachs@purdue.edu).}
% \thanks{Juan Wachs is with the School of Industrial Technology and the Department of Biomedical Engineering, Purdue University, West Lafayette, IN 47907 USA, and also with the Department of Surgery, Indiana University School of Medicine, Indianapolis, IN 46202 USA (e-mail: jpwachs@purdue.edu).}
\thanks{Richard Voyles is with the School of Engineering Technology, Purdue University, West Lafayette, IN 47907 USA (e-mail: rvoyles@purdue.edu)}
\thanks{*DR and MVB primarily conducted this research at Purdue University}}
% \thanks{$^{*}$Corresponding author is Deepak Raina}
%         <-this % stops a space

% \IEEEpubid{0000--0000/00\$00.00~\copyright~2023 IEEE}
\maketitle

\begin{abstract}
Ultrasound is widely employed for clinical intervention and diagnosis, due to its advantages of offering non-invasive, radiation-free, and real-time imaging. However, the accessibility of this dexterous procedure is limited due to the substantial training and expertise required of operators. {\color{black}The robotic ultrasound (RUS) offers a viable solution to address this limitation; nonetheless, achieving human-level proficiency remains challenging. Learning from demonstrations (LfD) methods have been explored in RUS, which learns the policy prior from a dataset of offline demonstrations to encode the mental model of the expert sonographer. However, active engagement of experts, i.e. Coaching, during the training of RUS has not been explored thus far. Coaching is known for enhancing efficiency and performance in human training. This paper proposes a coaching framework for RUS to amplify its performance}. The framework combines DRL (self-supervised practice) with sparse expert's feedback through coaching. The DRL employs an off-policy Soft Actor-Critic (SAC) network, with a reward based on image quality rating. The coaching by experts is modeled as a Partially Observable Markov Decision Process (POMDP), which updates the policy parameters based on the correction by the expert. The validation study on phantoms showed that coaching increases the learning rate by $25\%$ and the number of high-quality image acquisition by $74.5\%$.

\end{abstract}

% \vspace{-0.5cm}
\section{Introduction}
Medical ultrasound imaging is one of the most widely used imaging modalities for diagnostic and interventional procedures. Its widespread adoption can be attributed to its affordability, portability, non-invasiveness, absence of ionizing radiation, and real-time feedback. These merits make it particularly suitable for general use in low-income and developing regions. Unfortunately, the increasing demand for medical ultrasound, coupled with the substantial training required to become a proficient sonographer, and the significant impact of sonographer experience on diagnostic accuracy, contribute to a persistent gap in accessibility to this fundamental and valuable diagnostic tool \cite{carr2021influence}. 
% However, traditional ultrasound examinations require a substantial amount of clinical experience and hand-eye coordination skills for acquiring diagnostic quality images. 
% Inexperienced operators or novice sonographers may struggle to obtain accurate images, leading to non-diagnostic and misleading interpretations \cite{carr2021influence}. Since the experts are available in a limited capacity in rural areas or highly populated country, this led to delayed diagnosis and potentially poorer health outcomes.

% {\color{black}Focusing on coaching more. Training a robot to be an expert. Write it in a convincing way, reinforcing expertise.}
{\color{black}Robotic Ultrasound (RUS)} holds great promise to overcome these drawbacks. Further, RUS broadens the pool of skilled practitioners, enhances the safety of healthcare workers during pandemics, and improves the accessibility of ultrasound in rural areas where trained human practitioners are scarce \cite{raina2021comprehensive,chandrashekhara2022robotic}. A robot, once trained to an expert level, not only performs the procedure but also opens the possibility to train novice human operators. This dual functionality can significantly alleviate the training burden and offer a sustainable solution to the persistent gap in the availability of trained sonographers.
% Robotic Ultrasound System (RUS) holds great promise to overcome these drawbacks by offering reproducibility, improved dexterity, and intelligent decision-making on motion and imaging. But medical diagnostics are complex, highly variable with respect to the individual patient, and tightly tied to a high level of expertise. Accurate translation of such sophisticated skills to RUS has yielded a low likelihood of success with traditional programming approaches. 
% The training process includes showing novices the correct way to maneuver the probe and interpret the acquired images. In some cases, experts correct the novices' technique through hand-held guidance, which
Thus, transferring this skillset from an expert human operator to a robot is a key research question \cite{jiang2023robotic}. Novice sonographers learn this procedure by observing experts, followed by self-practice under the supervision of an expert, who may interrupt and provide corrections as needed. It brings into the picture the critical element of their training: coaching \cite{balakuntala2019extending}. It is defined as an active engagement of human experts in a trainee's learning process. 
%trim={L,B,R,T
\begin{figure}[t]
    \centering
    \includegraphics[trim=0cm 6cm 11cm 0cm, clip,width=\linewidth]{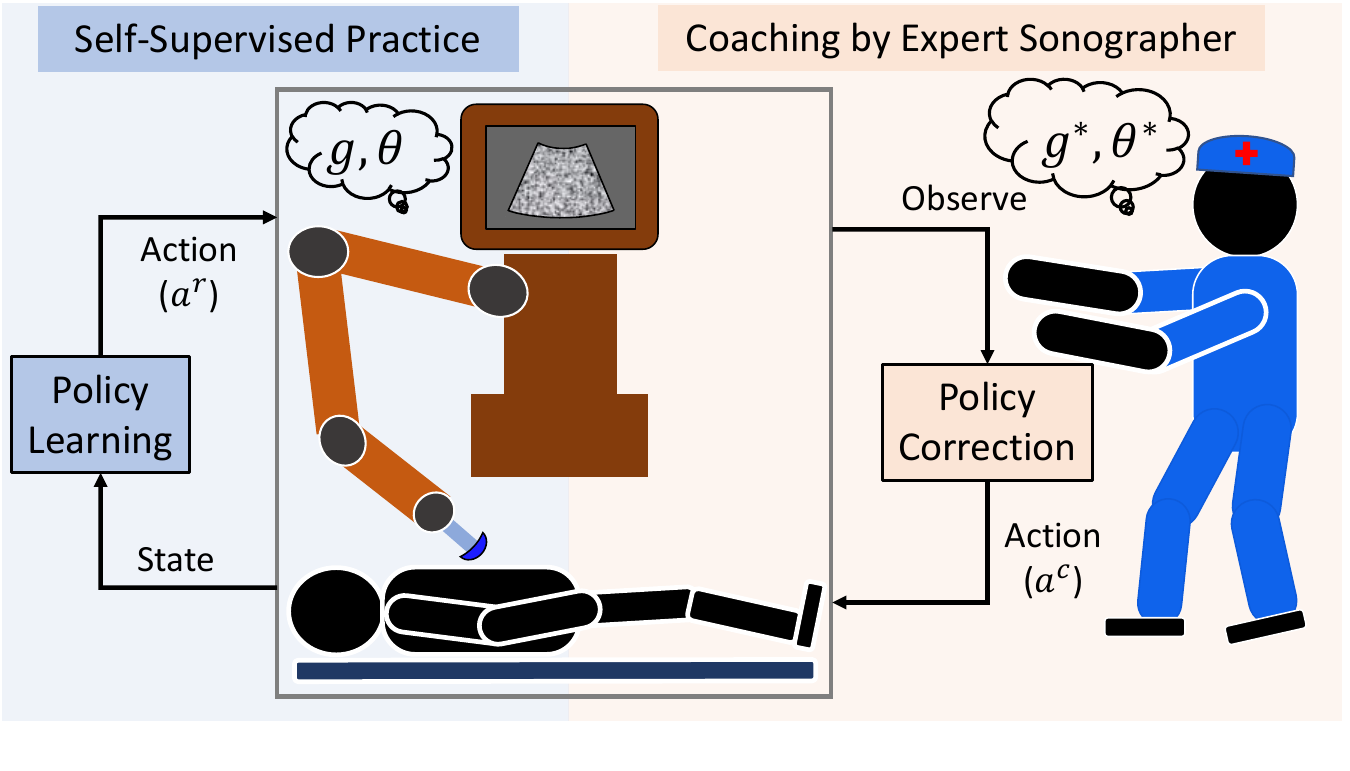}
    \caption{Improving the learning rate and performance of DRL policy for robotic ultrasound procedure with sparse coaching by expert sonographers.}
    \label{fig:coach_overview}
\end{figure}
% %trim={L,B,R,T
% \begin{figure}[t]
%     \centering
%     \includegraphics[trim=0cm 2cm 15cm 4.5cm, clip,width=\linewidth]{figs/teaser_coach}
%     \caption{Improving the performance of learnt DRL policy for robotic ultrasound procedure through coaching by experts}
%     \label{fig:coach_overview}
% \end{figure}

% {\color{black}not just RUS but in general robotics is lacking in coaching}
% When humans want to speed up learning and achieve very high levels of performance, they employ expert coaching\cite{haywood2021life}. 
Coaching by an expert guides novice humans to learn more rapidly and to achieve high levels of performance \cite{haywood2021life}. Surprisingly, there is a dearth of coaching research in RUS. Although, it has started gaining prominence in robot learning literature \cite{pamies2023autonomous, balakuntala2019extending}. {\color{black} For RUS, Deep reinforcement learning (DRL)
have been explored to mimic the training process of humans under reward-based self-supervision  \cite{li2021autonomous, ning2021autonomic}. However, these methods suffer from longer training times, local minima issues, and limited adaptability to unforeseen scenarios. Later, several works leveraged the knowledge of experts in the form of demonstrations and pre-trained a model offline \cite{raina2023expert, jiang2023intelligent}. These approaches are often termed as learning from demonstration (LfD). Notably, none of these methods considered real-time engagement of experts during training.} The aim of this paper is to develop a coaching framework for RUS to amplify its performance. There is a lack of formalism for coaching in robot learning. In this paper, we treat coaching as zero-shot learning because the RUS (learner) receives unlabeled corrective actions from the expert. These actions will correct the learnt policy and transform it towards an optimal policy.

% This modality is analogous to inverse optimal control methods where an agent infers
% an unknown objective by observing demonstrations [175], [176]. However, these methods
% 70
% demand complete task demonstrations to update the learned objective. Instead, we propose
% an online modification of the current objective via sparse local feedback. This approach is
% inspired by models for shard autonomy and physical human-robot interaction models [144]
% {\color{black}(Focus on the robot learning aspect, not computer vision) (Talk very briefly about traditional).} 
\subsection{Related work}
\subsubsection{Robotic ultrasound systems}
Earlier RUS used visual servoing to automate the ultrasound procedure \cite{mebarki20102, nadeau2016moments}, but these methods lack human-level anatomical knowledge, a crucial factor for sonographers in relating probe motions to the complex anatomy. To supplement this knowledge, later methods used high-end imaging modalities like MRI, CT, or 3D data  \cite{hennersperger2016towards, al2021autonomous} to identify the anatomical landmarks and pre-plan the probe trajectory. However, the sparse availability of these modalities in underserved regions did not solve the underlying issue. Recent works have explored the use of DRL architectures to mimic the self-supervised practice of human trainees \cite{li2021autonomous, ning2021autonomic}, yet the clinical applicability of these systems was questioned due to longer training times. 

% Early RUSs used traditional computer vision techniques to fully automate the technique. Visual servoing-based approaches extracted image features, such as moment-based \cite{mebarki20102}  or three-dimensional \cite{nadeau2016moments}, and then mapped them to the probe’s motion using a pre-conceived control law. (But it does not capture the expertise.)
% However, these techniques had limitations in representing complex human anatomies through hand-engineered features, mapping limited feature knowledge to achieve comprehensive control of probe, addressing occlusion of structures in images, and avoiding calibration errors. Over the years, researchers also made efforts to leverage imaging modalities like Magnetic Resonance Imaging (MRI), Computed Tomography (CT) (However, they do not require experience), and 3D patient data for localizing the region of interest and pre-planning the scanning trajectory \cite{hennersperger2016towards}. These approaches were limited by the inaccessibility of high-end imaging modalities in rural or underserved regions and the computational expenses associated with their real-time analysis.

% (Key point: Coaching has not been explored in robotic learning and RUS is)

% Lately, researchers have explored the use of Deep Reinforcement Learning (DRL) architectures to mimic the self-supervised practice of human trainees \cite{li2021autonomous, ning2021autonomic}. 
\subsubsection{Experts' engagement in learning of RUS}
{\color{black}Researchers explored the engagement of ultrasound experts through LfD approaches to reduce training times. 
% \cite{janvier2008performance} first attempted to teach ultrasound skills to a prototype robotic arm. However, the system simply replayed the demonstrated position and force trajectories. Thus, it can only be replicated in a similar environment. 
Lonas et al. \cite{mylonas2013autonomous} used the Gaussian Mixture Model (GMM) and Gaussian Mixture Regression (GMR) to learn a probe motion model from offline demonstrations. However, the model did not include the probe forces and ultrasound image information, which limited its clinical applicability. Li et al. \cite{li2022learning} used the dataset curated from experts' demonstrations to sample probe poses during model-free RL. 
% A task-quality model was also learned from expert demonstration, which was used during RL to either select predicted poses/forces or sample from the expert's dataset. 
This approach has two demerits. First, a very large dataset would be required to handle human anatomical variability. Second, sampling from a large dataset would be computationally expensive, hence impractical for RUS. Raina et al. \cite{raina2023robotic, raina2023deep, raina2023rusopt} proposed to model the Gaussian process (GP) prior and kernel from offline expert demonstrations. Later, this pre-trained GP was used in the Bayesian optimization framework for the robotic acquisition of optimal images.
Jiang {et al.} \cite{jiang2023intelligent} and Burke et al. \cite{burke2023learning} inferred the reward for optimizing the RUS policy from scanning demonstrations, which assumed that the images shown in the later stage are more important than the earlier images. It is important to note that the above-cited works proposed to learn offline using a dataset collected from expert demonstrations, which often require numerous optimal demonstrations. Additionally, these approaches are goal-driven, while the proposed coaching framework allows updating the policy objectives and parameters through local corrections to the robot's trajectory.} 
{\color{black}
\subsubsection{Coaching robots}
% Most LfD methods require numerous successful demonstrations for task learning. In addition, the major drawback of LfD is goal-driven learning. 
% Recent extensions to LfD have considered incorporating interaction during the transfer process \cite{billard2016learning}. 
Early coaching-based methods incorporated diverse human feedback within reinforcement learning methods \cite{thomaz2005real}. Macglashan et al. \cite{macglashan2017interactive} proposed COACH (Convergent Actor-Critic by humans) that extended RL with online reward/penalty from experts. However, COACH converged only to local minima and relied on sparse binary evaluations using goal examples.  
% However, COACH converged only to local minima, did not require regular evaluation, and relied on sparse binary evaluations using goal examples. 
As such, this method demands substantial sampling, greater learning time, and a large feedback volume. Later, preference-based learning techniques were introduced into DRL, which involved presenting action samples (trajectories) to experts for grading and later enhancing behavior based on their preferences \cite{sadigh2017active, biyik2018batch}. 
% However, these methods lacked clarity on when and how to generate effective trajectories, leading to data inefficiency. Active learning approaches addressed this by selecting informative trajectories through metrics like volume removal, information gain, or mutual information optimization \cite{basu2019active, billard2016learning}. 
While preferences aid improvement, the challenge remained in estimating trajectories to elicit preference-based feedback, resulting in inefficient use of expert's input. Expert feedback is not only evaluation but also provides useful information on direct trajectory modifications \cite{griffith2013policy}. Therefore, the robot should leverage these corrections to optimize actions and update rewards.

Recent strategies in physical human-robot interaction (pHRI) considered corrections as informative rather than disturbance \cite{losey2019learning}. Online learning from correction aimed to refine objectives and generate the trajectory based on comparisons between corrected and original trajectories \cite{bobu2020quantifying, li2021learning}. Similar techniques have been employed in shared autonomy in manipulation \cite{javdani2018shared}. 
% In contrast, offline correction-based learning iteratively enhanced underlying objectives for subsequent task iterations \cite{jain2015learning}. These methods, akin to human coaching, observed corrections as deliberate and valuable. 
However, the challenge lies in the robot understanding the context of interactive feedback and updating its objectives and policy accordingly. A formalism for coaching that combines policy updates and learning new objectives is lacking in the robotics literature. In order to address these challenges, we extend Partially Observable Markov Decision Process (POMDP) formalizations from pHRI and shared autonomy to coaching the DRL policy. Moreover, we demonstrate its applicability to previously unexplored and highly expert-dependent modality of medical ultrasound imaging.}

% \vspace{-2mm}
\section{Methodology}
{\color{black}The pipeline of the methodology is outlined in Algorithm 1}, which combines self-supervised practice through DRL with coaching through sparse expert feedback. The robot learns a DRL policy to perform ultrasound under self-supervision. During learning, the expert provides online feedback through kinesthetic corrections, which will update the policy objective and parameters towards optimal policy.
% \vspace{-0.3cm}
\begin{algorithm}[h]
\SetAlgoLined
% \textbf{Input:} Prior $\mathcal{E}{(\boldsymbol{\theta})}$, Region $A$, max. iterations $N_{max}$\;
Initialize DRL policy network $\pi_\theta(s)$, replay buffer $\mathcal{R}$, and coach replay buffer $\mathcal{R}^c$\;
  \For{$i = 1,...,N_{max}$ }{
    \For{each step}{
    Sample $a_t$ from $\pi_\theta(a|s)$ and execute $a_t$\;
    Observe new state $s_{t+1}$\;
    Store ($s_t, a_t, r(s_t, a_t), s_{t+1}$) in $\mathcal{R}$\;}
    % \For{each DRL gradient step}{
    % Sample from $\mathcal{R}$ and update parameters $\theta$\;
    % }
    \If{ coaching correction $a_t^c$}{
    Compute corrected trajectory $\Pi^c$;\\
    Store ($s_t, a^c_t, r(s_t, a^c_t), s_{t+1}$) from $\Pi^c$ in $\mathcal{R}^c$\;
    }
    \For{each coaching gradient step}{
    Sample from $\mathcal{R}^c$;\\
    Update parameters $\theta$\ using KL divergence;
    % $\mathcal{L}_{KL} = \mathbf{D}_{KL}(\pi_\theta(a|s)~||~{\pi}(a|s, \hat{\theta}^*))$
    }
    % $\boldsymbol{p}_i \leftarrow \arg\max_{\boldsymbol{p} \in A} EI(\boldsymbol{p})$\;
    % \eIf{termination criteria reached}{
    %     stop\;
    % }{
    %     Probe at~$\boldsymbol{p}_i$, compute image quality $q_i$\;
    %     Set $\Bar{\mathbf{p}} \gets \Bar{\mathbf{p}} \cup \{p_i\}$, $\Bar{\mathbf{q}} \gets \Bar{\mathbf{q}} \cup \{q_i\}$\;
    %     $\boldsymbol{\theta} \gets \text{argmax} \, \mathcal{L}(\boldsymbol{\theta}|\Bar{\mathbf{p}},\Bar{\mathbf{q}}) \mathcal{E}(\boldsymbol{\theta})$\; 
    %     Set $\Bar{\mathbf{f}} \gets \Bar{\mathbf{f}} \cup \{ q_i - {\mu}_{\boldsymbol{\theta}}(\boldsymbol{p})\}$\;
    %     Re-estimate GP\;
    }
 % \Return Top probe poses with max. image quality\;
 \caption{\color{black}Coaching a RUS system}
 \label{alg:BO}
\end{algorithm}
% \vspace{-0.7cm}
\subsection{Self-supervised practice through DRL}
% This section describes the RL policy,
% which is learned based on an image quality reward. The robot model executes in a finite bounded horizon, with a Markov decision process $\mathcal{M}$, with state space $\mathcal{S}$ and action space $\mathcal{A}$. For a horizon $T$, the state transitions according to the dynamics $\mathcal{T}:\mathcal{S} \times \mathcal{A} \longrightarrow \mathcal{S}$.
% \subsubsection{Policy}
This section describes a self-supervised practice through DRL policy, which is learnt using a reward formulation based on ultrasound image quality. The robot model executes in a finite bounded horizon, with a Markov decision process $\mathcal{M}$, with state space $\mathcal{S}$ and action space $\mathcal{A}$. For a horizon $T$, the state transitions according to the dynamics $\mathcal{T}:\mathcal{S} \times \mathcal{A} \longrightarrow \mathcal{S}$. A policy $\pi_\theta (a|s)$ represents the probability of taking action $a$ given a state $s$, with parameters $\theta$. The cumulative expected reward over horizon $T$ is given by eq. \eqref{eq:1}. 
\begin{equation}
    J(\pi) = \mathbb{E}_{\pi} \left[\sum_{t=1}^{T} \phi(s_{t},a_{t})\right]
\label{eq:1}
\end{equation}
We used an off-policy algorithm, Soft Actor-Critic (SAC) \cite{haarnoja2018soft} to learn the policy. This is due to the better sample efficiency and generation of stable policies by SAC, which is suitable for practical
applications such as manipulation \cite{singh2019end}. This algorithm maximizes the cumulative reward and entropy to learn the policy. 
% The state space, action space and rewards of this policy are explained below.
\subsubsection{State space}
\begin{figure}[!ht]
	\centering
	%trim={L,B,R,T}
	\includegraphics[trim=0cm 6.4cm 9cm 0.2cm,clip,width=\linewidth]{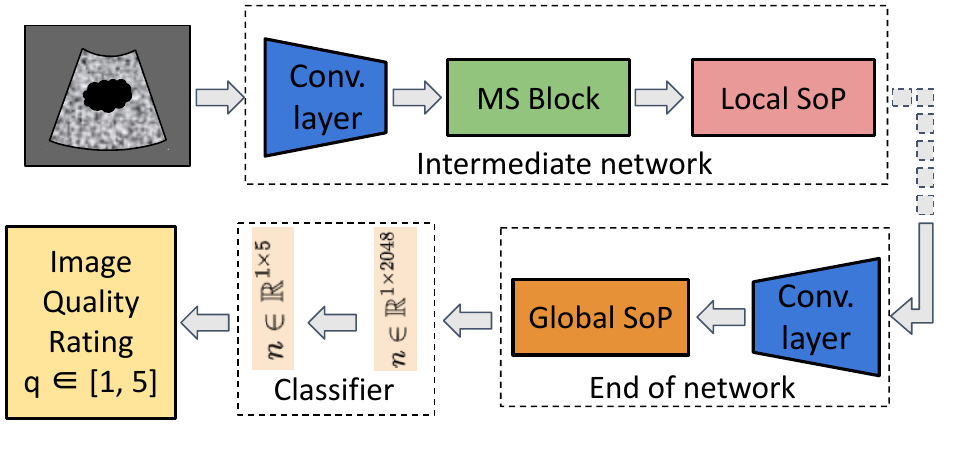}
     % \vspace{-0.8cm}
	\caption{\color{black}State space representation using a deep convolution neural network}
	\label{fig:quality_network}
\end{figure}

The state $\mathcal{S}$ is defined based on the ultrasound image. {\color{black}We have adopted an image quality classification network from our previous work \cite{raina2024deep}, which used ResNet50 as a base network with multi-scale and higher-order processing of the image for conducting the holistic assessment of the image quality.} The block diagram of this network is shown in Fig. \ref{fig:quality_network}. 
% \vspace{-0.2cm}
\vspace{-0.2cm}
{\color{black}This classifier first extracts features at multiple scales to encode the inter-patient anatomical variations. Then, it uses second-order pooling (SoP) in the intermediate layers (local) and at the end of the network (global) to exploit the second-order
statistical dependency of features. The local-to-global SoP will capture the higher-order relationships between different spatial locations and provide the seed for correlating local patches.}
This network encodes the image into a feature vector of size $2048$, which represents the state of the policy. 
\subsubsection{Action space}
The action space $\mathcal{A}$ for the SAC policy is a combined position, orientation, and forces of the probe. Specifically, the robot controls the position of the probe in the $xy-$plane; orientation along $roll$, $pitch$ and $yaw$; and force along the $z-$axis (normal to the surface). 
% Thus, it localizes the feature and applies appropriate force to obtain good-quality images.
\subsubsection{Rewards}
The reward is based on the ultrasound image quality estimated using the same network \cite{raina2024deep}, which represents the state. {\color{black}The extracted image features are passed through a linear classifier layer to generate a feature vector of size $5$. Finally, the index of maximum value for this feature vector gives an integer quality rating between $1-5$}. In addition, a quality rating of $0$ is assigned when the measured force value of the probe along the z-axis is below the minimum required for the appropriate contact. The reward is then defined as
% \vspace{-0.2cm}
\begin{equation}
    r_u = \eta (q==q_{max}) + q/q_{max} \label{eq:rew_u}
\end{equation}
% \vspace{-0.cm}
where $q$ is the quality of the image, and the expression $(q==q_{max})$ is $1$ if the quality is maximum and $0$ otherwise. The constant $\eta$ is used to amplify the reward when the robot reaches the maximum image quality (i.e., $q=q_{max}$). This image quality-based reward guides the self-supervised learning of the ultrasound policy.
% \vspace{-0.3cm}
\subsection{Coaching with sparse expert's feedback}
The coaching scenario for RUS is shown in Fig. \ref{fig:coach_overview}. It is treated as learning a hidden goal ($g^*$) by observing the corrective actions ($a^c$) provided by the expert. We develop a formalism for representing coaching as a partially observable dynamical system. Coaching aims to improve the objectives and parameters through local corrections to the trajectory. 
% The coaching model will allow the robot to reason over objectives by observing the corrections. 
% The state at any time t is represented as $s_t$, $a_t^r$ be the robot’s action. The feedback provided by the human is a desired action, represented as $a_t^h$. 
% Let the robot transition to next state $s^'$ based on transition dynamics  $\mathcal{T}(s^'|s,a^r+a^h)$.
RUS can only observe the coach's corrections $a^c$ and its actions $a^r$. It acts according to its optimal policy ($\pi_\theta$), however, the coach expects it to operate with respect to a true objective whose optimal actions are determined by $\pi_{\theta^*}$. The coach does not directly provide parameters $\theta^*$, nor does the RUS know $\pi_{\theta^*}$.
{\color{black}In DRL policy learning, we assumed that the goal states ($g$) are known and the reward is computed based on eq. (2)}. However, a correction from the coach implies that the goals and the policy need to be updated. If the current policy parameters $\theta = \theta^*$, then the formulation is an MDP where the robot is already behaving optimally with respect to expectations. However, when $\theta \neq \theta^*$, the robot cannot directly observe $\theta^*$ to update its policy. Further, the robot cannot observe the coach's expected goals $g^*$ as well.  Therefore, the uncertainty in the objectives $g^*$ and corresponding policy parameters $\theta^*$ turns this into a Partially Observable Markov Decision Process (POMDP).
\subsubsection{POMDP model}In this POMDP, $g^*$ forms the hidden part of the state $\bar{s}$, and the coach's corrective actions $a^c$ are observations about $g^*$ and $\theta^*$ under some observation model as $\mathcal{O}(a^c | \bar{s}^* = (s, g^*), a^r)$. The observation of the coach's correction allows the robot to learn the true objective. The coach's feedback is modeled as corrections that optimize the expected return from the state $\bar{s}$ while taking action $a^r+a^c$. The action-value function ($Q$) captures this expected return. Thus, it can be written as:
\begin{equation} \label{eq:q-value-function}
    \mathcal{O}(a^c|\bar{s},a^r) \propto e^{Q_{\pi_\theta}(\bar{s},a^r+a^c)}
\end{equation}

The relationship for the observation model indicates that the coach provides feedback,
which together with the robot’s action, will lead to the desired behavior. And similar to
human coaching, the robot is expected to continuously learn a better objective by observing
the feedback. This formal approach captures the true essence of human coaching. The uncertainty is in the estimate of the desired goals $g^*$ and the corresponding policy parameters $\theta^*$.
The environmental state $s$, which is part of the POMDP state $\bar{s}$, is assumed to be fully observable, similar to POMDP-lite \cite{chen2016pomdp}. However, the robot cannot observe the goal part of the state $\bar{s}$. Instead, the robot can learn an action-value function over belief states $b(\bar{s})$ as follows:
\begin{equation}
    Q^{*}(b, a^c+a^r) = \mathbb{E}[r(b, a^c + a^r) + \mathbb{E}_b^{'}[V*(b')]]
\end{equation}
Note that solving a POMDP with continuous action and states is expensive and usually intractable. Several approaches to estimate and approximate solutions for a POMDP have been explored in literature, such as hindsight optimization \cite{javdani2018shared} or reduction to QMDP \cite{losey2022physical}. We provide an approximation to simultaneously update the policy parameters $\theta$ and the action value at the corrected states.

\subsubsection{Approximate solution}
\begin{figure}[ht]
    	\centering
    	%trim={L,B,R,T}
    	\includegraphics[trim=0cm 11.5cm 19cm 0cm,clip,width=0.8\linewidth]{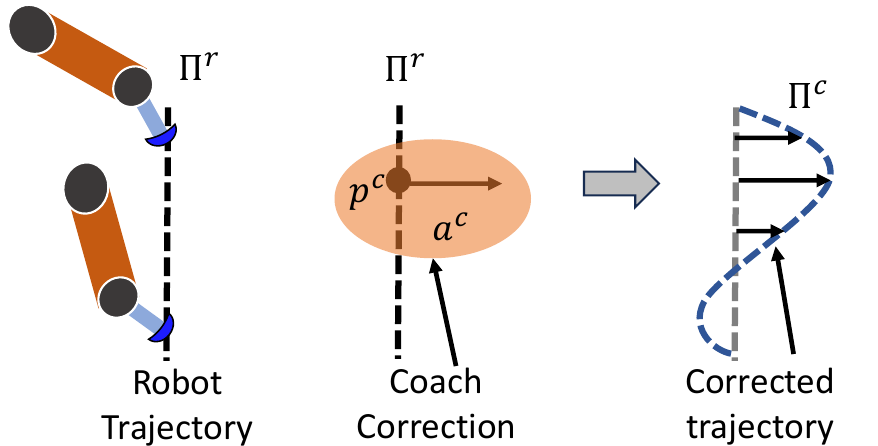}
    	\caption{\color{black}The correction of trajectory $\Pi^r$ based on coach's correction to obtain preferred trajectory $\Pi^c$}
    	\label{fig:traj_correction}
    \end{figure}

In order to solve the POMDP, we transform it into a policy update problem. Several approximation are defined to achieve the policy update, which include trajectory correction, reward formulation, and policy parameter computation. These approximations provide an elegant solution to the POMDP model.
\\
\\
% These approximations are explained below:
% \begin{itemize}
%     \item \textbf{Trajectory correction :} 
    \textbf{Trajectory correction :}     {\color{black}The Q-value shown in eq. \eqref{eq:q-value-function} cannot be computed for continuous state and action spaces. Therefore, the reasoning is done in the trajectory space instead of the control action space. The trajectory space is defined based on pose and force profiles resulting from the DRL policy learning. First, the robot estimates a trajectory based on the learned policy $\pi_\theta(a|s)$ derived from the observed goal $g$. The robot then utilizes an in-built hybrid force-position controller to track this trajectory. Instead of computing the Q-value for each action, we can estimate the total reward resulting from following this trajectory $\Pi^r$. 
    Fig. \ref{fig:traj_correction} shows an example of a scenario where the robot is following the trajectory determined by the DRL policy. The robot trajectory can be represented as a sequence of control inputs resulting from the actions generated by $\pi_\theta$.}
    \vspace{-0.5cm}
    % \vspace{-0.3cm}
    The coach can apply a correction at any instant after some duration of DRL policy learning. This corrective action $a^c$ is provided at a point $p^c$ on the robot trajectory. {\color{black}However, the point correction has to be propagated across a local region of the trajectory. A trajectory optimizer is used to obtain the coach-preferred trajectory}. The coach-preferred trajectory ($\Pi^c$) is obtained by smoothly deforming the robot trajectory ($\Pi^r$) locally using minimum jerk constraints, as follows:
    \begin{equation}
        \Pi^c = \Pi^r + \mu\Pi^{o}(a^c) 
    \end{equation}
    where $\mu>0$ is a scaling factor for the deformation and $\Pi^{o}$ is offset to current trajectory based on action $a^c$ computed using minimum jerk optimization. {\color{black}The minimum jerk trajectory is computed by obtaining a trajectory $\Pi^{o}$ that minimizes the integral of a squared jerk over time. The solution to this optimization is a trajectory represented by a quintic polynomial as $\Pi^{o}_t = \sum_{i=0}^5k_i t^i$. Here, $t$ is the time, and $k_i$ are coefficients of the polynomial to be determined based on control points and boundary conditions. The first three constants can be determined from the initial position, velocity, and acceleration at $t = 0$. Similarly, the last three can be estimated based on the final or target position, velocity, and acceleration. For the case of offsetting the robot trajectory $\Pi^r$, we generate piecewise minimum jerk trajectory with control points at a $p^c$ and two points on either side of $p^c$ spaced at a time distance determined by the scale of the correction.} Once the trajectory is updated, we can run the robot along the new trajectory and discover the associated states observed while moving on the expert-preferred trajectory.
    % \item \textbf{Reward modification:}
    \\
    \\
    \textbf{Reward modification:}
   {\color{black}The coach-preferred trajectory is only a trajectory offset based on the corrective action. The robot has not learned how to leverage the knowledge from the preferred trajectory to improve the policy.} We propose augmenting the reward objectives with two components as coach reward $r_{c}(\bar{s}, a)$ and trajectory reward $r_{\Pi} (\bar{s},a)$, as follows:
    \begin{equation}
        r(\bar{s},a) = w_u r_u(\bar{s},a) + w_c r_c(\bar{s},a) + w_\Pi r_\Pi(\bar{s},a)
    \end{equation}
    where $w_*$ is the weight associated with corresponding rewards ($r_*$). The coach reward $r_c$ associates the states observed by following the coach-preferred trajectory $\Pi^c$ with a small positive reward. The trajectory reward $r_\pi$ penalizes large changes in poses or forces to ensure that the policy generates a smoother path. {\color{black}The reward weights $w_*$ are learned by maximizing the return to choose the preferred trajectory over the old robot trajectory}
    % \item \textbf{Policy update:}
    \\
    \\
    \textbf{Policy update:}
    In addition to updating the reward or goals, the policy parameters are offset to move towards the coach’s optimal policy $\pi_{\theta^*}$. Several iterations of coaching will generate multiple desired states and segments of preferred trajectories. The state transitions, the corresponding rewards from the new reward model, and the actions along these trajectories are stored in a separate replay buffer, named as \textit{coach replay buffer}. Then, an approximation of the desired optimal policy is computed.  The approximate policy $\pi_{\hat{\theta}}{(a|s)}$ generates the actions needed to move along the preferred trajectories. The parameters of the policy are represented as $\hat{\theta}^*$ because they are an approximation of the true optimal parameters $\theta^*$.   The policy can be written as
    $\pi_{\hat{\theta}^*}(a|s) \rightarrow \pi(a|s, \hat{\theta}^*)$. A Gaussian distribution is used for the approximate policy, i.e., $\pi(a|s, \hat{\theta}^*) \sim \mathcal{N}(\mu(\hat{\theta}^*), \sigma(\hat{\theta}^*))$. We know the sequence of states $s$ on the expert-preferred trajectories and the corresponding action to take in these states. Therefore, we aim to estimate the policy parameters $\hat{\theta}^*$ that fit this state and action data. The parameters are estimated using maximum likelihood estimation. Once the approximate policy is found, the current policy is regularly updated by performing gradient steps of SAC on the coach replay buffer. For the policy update during SAC, we include an additional loss based on KL divergence between the robot’s policy $\pi_\theta(a|s)$ and the approximation $\pi(a|s,\hat{\theta}^*)$.
    \begin{equation}
        \mathcal{L}_{KL} = \mathbf{D}_{KL}(\pi_\theta(a|s)~||~{\pi}(a|s, \hat{\theta}^*))
    \end{equation}
    This KL divergence loss ensures actions are taken according to the coach's preferences in the preferred states. 
\section{Results and Discussions}
\subsection{Experimental setup}
\begin{figure}[t]
	\centering
	%trim={L,B,R,T}
	\includegraphics[trim=0cm 0.58cm 0cm 0cm,clip,width=\linewidth]{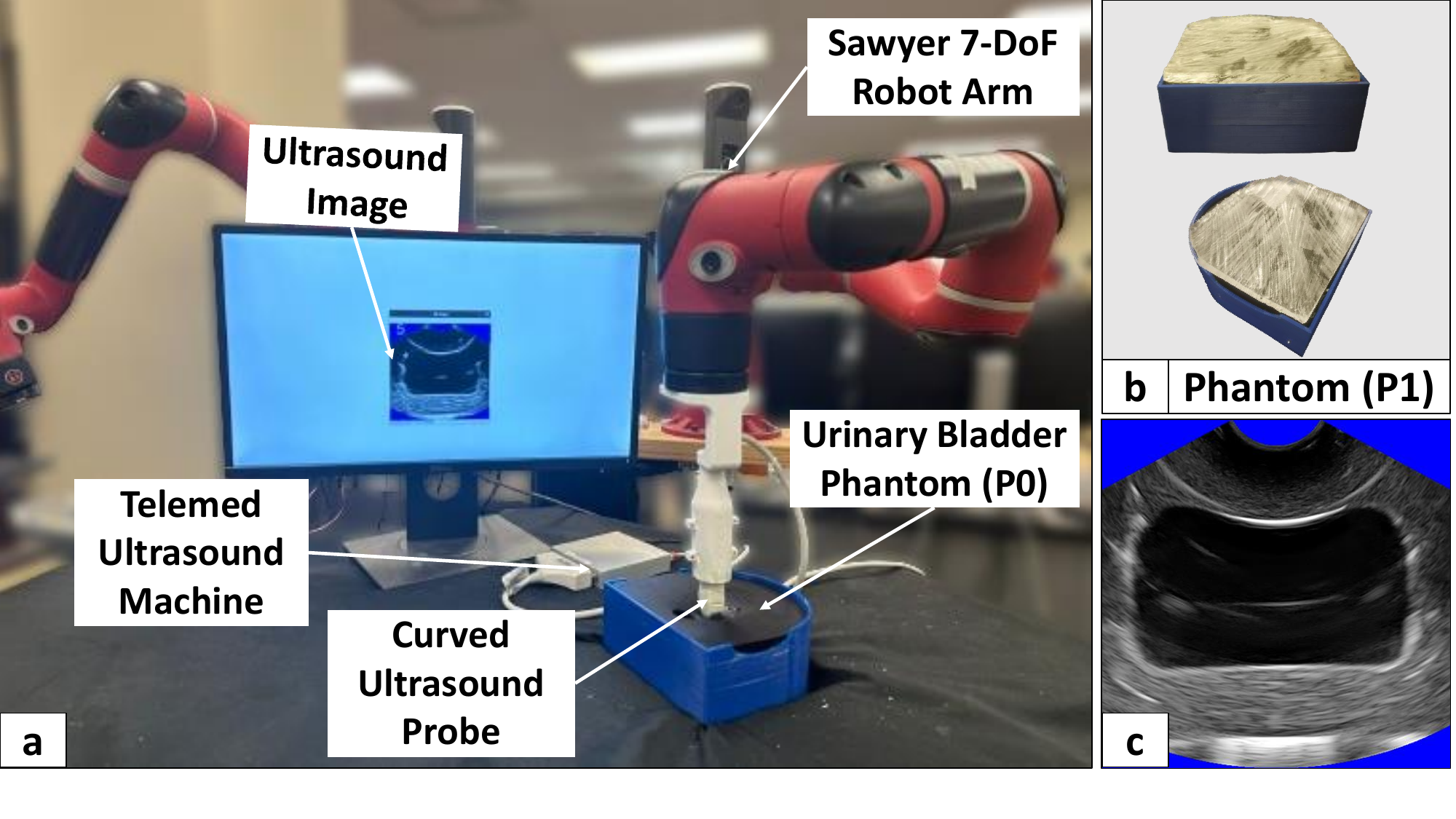}
     % \vspace{-0.8cm}
	\caption{(a) Robotic ultrasound system (b) Urinary bladder phantom (P1) for testing (c) Acquired high-quality ultrasound image ($q=5$) from phantom P0}
	\label{fig:exp_setup}
\end{figure}

The experimental setup is shown in Fig \ref{fig:exp_setup}. It consists of a 7-DOF Rethink Robotics Sawyer arm, with a Micro Convex MC10-5R10S-3 probe by Telemed Medical Systems, USA, attached to its end-effector using a custom-designed gripper. {\color{black}The robot has a wrist-mounted force-torque sensor, which was used to measure the forces.} Two urinary bladder phantoms, P0 and P1, were used for scanning. P1 is a modified variant of P0 with a ballistic gel layer. ROS was used as the middleware to transmit images and commands across the devices.
% \vspace{-0.7mm}
% \vspace{-0.4cm}
\subsection{Implementation details}
To analyze the effect of coaching on learning performance, we conducted experiments by inducing coaching corrections at different iterations of training.
Four policies were trained for the following cases: (i) No Coaching; Coached after every (ii) $20k$, (iii) $10k$, (iv) $5k$ timesteps. {\color{black} The policy with no coaching was based on reward in eq. (2) and policies with coaching were based on a combined reward in eq. (6).} Each policy was trained for a maximum of $200k$ steps or until convergence, {\color{black}approximately equal to eight hours of wall clock time. The robot was initialized to a random pose for each episode of training. An episode of training ends upon either achieving the high-quality image or reaching the $50$ steps.} The value of $\eta$ in eq. \eqref{eq:rew_u} is empirically set to 10 to maximize performance. The action space limits are set as $x \in (-0.05, 0.05)$m, $y \in (-0.03, 0.03)$m, $fz \in (5 - 30)$N, $roll \in (-0.2, 0.2)$rad, $pitch \in (-0.2, 0.2)$rad,
and $yaw \in (-0.5, 0.5)$rad. {\color{black}The roll, pitch and yaw angles were carefully selected to ensure collision-free scanning of the phantoms. The hybrid position-force control mode of Sawyer was used to control the robot.}

The coaching corrections are provided by human experts through kinesthetic interactions with the robotic arm. {\color{black}At the instant of interaction, the robot movement is paused, which was hardcoded in the training script. Then, the expert activated the free-drive mode of the robot by pressing the button provided on the wrist and nudged the robot toward the optimal trajectory. Once the expert finished the coaching, which is indicated by a high-quality image ($q\geq4$), the training script was re-initialized from the same time step. Note that the policy weights were updated before re-initialization.} 
% This switching control method effectively handled the discontinuity between the contact and free motion. }

% {\color{black}During the training process, if the robot is at a step requiring coaching correction, the expert will intervene by utilizing the free-drive mode button located on the wrist. This free-drive mode will allow the expert to free move the arm. Once the necessary correction will be made, the robot will resume its training from the exact time step where it was halted.} 
% \vspace{-0.2cm}
\subsection{Performance evaluation} 
\begin{figure}[!ht]
	\centering
	%trim={L,B,R,T}
    % \includegraphics[trim=0cm 0.2cm 0cm 0cm,clip,width=0.9\linewidth]{figs/coaching_plots_final_v1}
    \includegraphics[trim=0cm 0.2cm 0cm 0cm,clip,width=0.9\linewidth]{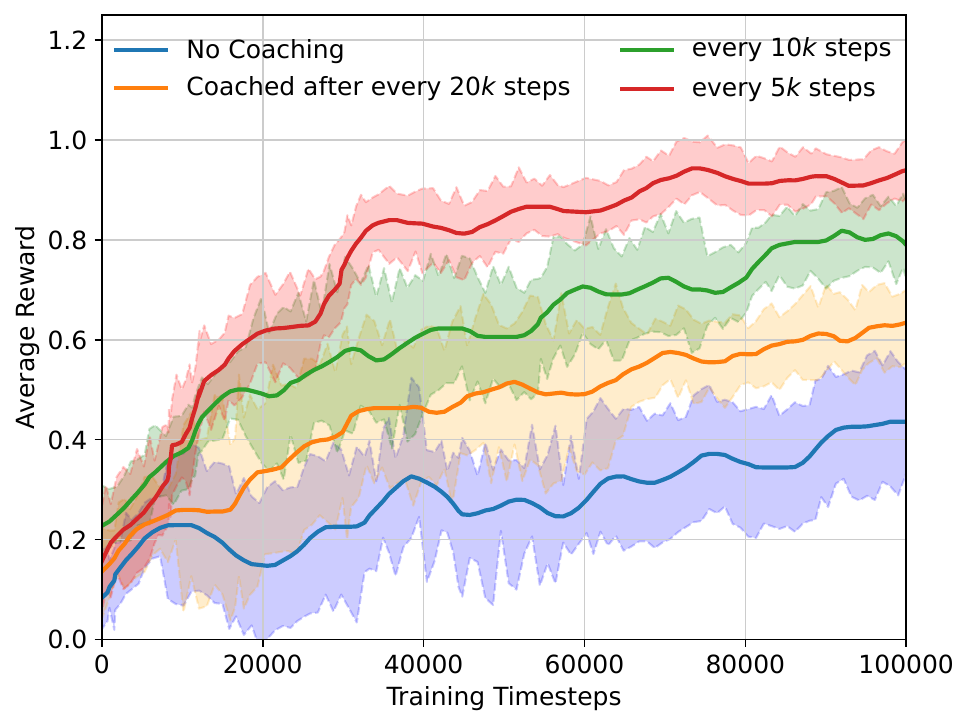}
    \caption{Comparison of average reward over training timesteps for policy learning with and without coaching}
	\label{fig:avg_reward}
\end{figure}

 Fig. \ref{fig:avg_reward} compares the normalized average reward over the training timesteps for different policies. One of the primary issues associated with the ``no coaching" policy was the tendency to explore sub-optimal states (i.e., low image quality regions). The poses and forces necessary for producing high-quality images constitute a narrow domain within the action space. Consequently, the ``no coaching" policy often chooses inappropriate actions (probe poses and force). To correct this behavior, the coach intervened by adjusting either the probe position or orientation relative to the phantom surface, or by modifying the force applied to the phantom. After coaching, the reward from the resulting policy demonstrated improvement at each step of the training, wherever it was introduced. Coaching also improved the learning efficiency, as seen by the faster convergence of coached policies in Fig. \ref{fig:avg_reward}. In the absence of coaching, the policy failed to approach the optimal region even after $100k$ steps and showed no convergence even at $200k$ steps. In contrast, policies with coaching introduced at intervals of $20k$ and $10k$ timesteps demonstrated convergence at approximately  $150k$ ($\downarrow\sim25\%$) and $120k$ ($\downarrow\sim40\%$) steps, respectively. The policy receiving maximum coaching (every $5k$ steps) converged at $100k$ steps. These findings underscore the efficacy of coaching in identifying optimal policies at a faster learning rate.

Once the policies were trained, we compared the performance of learned policies during execution. Each policy is executed for $10$ trials, and each trial has a maximum of $50$ steps. The following metrics are used for comparison: (i) Number of High-Quality Images (HQI) sampled ($q\geq4$), (ii) First instance of HQI sampling, (iii) Errors in probe motion. The metric (iii) measures the error between probe position ($\boldsymbol{p}$), orientation ($\boldsymbol{o}$) and force ($f_z$) at the step with the best reward and their corresponding Ground Truth (GT) values. The evaluated mean $\pm$ standard deviation values of these metrics are given in Table \ref{tab:comparison}.
\setlength{\tabcolsep}{4pt}
\renewcommand{\arraystretch}{1.15}
{\begin{table}[!ht]
    \centering
\caption{\color{black}Testing performance of trained policies on phantoms P0 and P1}
\resizebox{\linewidth}{!}{\begin{tabular}{cc|cccccccc}  
    \toprule
    ~ &  \multicolumn{2}{c}{\multirow{2}{*}{\textbf{Policy}}} &  \textbf{No. of} & \textbf{First} & \multicolumn{3}{c}{\textbf{Error in probe motion}} \\
    \cline{6-8}
    ~ & \multicolumn{2}{c}{} &\textbf{HQI} & \textbf{HQI step} & $\boldsymbol{p}$ (m) $\times10^{-1}$ & $\boldsymbol{o}$ (rad.) &$f_z$ (N)   \\
    \midrule 
    \multicolumn{1}{c|}{\multirow{4}{*}{\rotatebox[origin=c]{0}{\textbf{P0}}}} & \multicolumn{2}{c}{\textbf{No Coaching}} & $2.1\pm 0.5$ & $45.0\pm 1.1$ & $0.32\pm 0.10$ & $0.23\pm0.12$ & $8.2\pm 1.1$ \\
    \multicolumn{1}{c|}{} & \textbf{Coached} & $n=20k$& $8.2\pm 4.6$  & $35.3\pm 8.1$ & $0.23\pm 0.08$ & $0.15\pm0.09$ & $3.5\pm 1.8$ \\
    \multicolumn{1}{c|}{} & \textbf{after every} & $n=10k$ & $15.3\pm 4.1$  & $19.8\pm 4.0$ & $0.11\pm 0.06$ & $0.11\pm0.07$ & $2.5\pm 0.8$ \\
    \multicolumn{1}{c|}{} & $n$ \textbf{steps} & $n=5k$ & $21.5\pm 2.3$  & $10.2\pm 1.5$ & $0.05\pm 0.02$ & $0.06\pm0.01$ & $0.7\pm 0.3$ \\
    \midrule
    \multicolumn{1}{c|}{\multirow{4}{*}{\rotatebox[origin=c]{0}{\textbf{P1}}}} & \multicolumn{2}{c}{\textbf{No Coaching}} & $1.5\pm 0.3$ & $47.0\pm 0.8$ & $0.38\pm 0.09$ & $0.30\pm0.14$ &  $10.0\pm 1.8$ \\
    \multicolumn{1}{c|}{} & \textbf{Coached} & $n=20k$& $5.9\pm 1.6$  & $38.1\pm 4.2$ & $0.26\pm0.05$ & $0.22\pm0.16$ & $4.1\pm 1.7$ \\
    \multicolumn{1}{c|}{} & \textbf{after every} & $n=10k$ & $12.3\pm 2.4$  & $25.2\pm 5.9$ & $0.17\pm 0.05$ & $0.17\pm0.11$ & $3.1\pm 0.6$ \\
    \multicolumn{1}{c|}{} & $n$ \textbf{steps} & $n=5k$ & $16.2\pm 2.9$  & $12.1\pm 3.5$ & $0.09\pm 0.05$ & $0.10\pm0.03$ & $0.9\pm 0.2$ \\
    %Every step & & & & \\
    % Random  &   &  &  &  \\
    \bottomrule
    \end{tabular}}
    \label{tab:comparison}
\end{table}}

For Phantom P0, the results showed an improvement in HQI from $2.1$ to $8.2$ ($\uparrow74.4\%$) with coaching. Specifically, for the test phantom P1, the ``no coaching" policy resulted in a low HQI of $1.5$. However, the policy with coaching after every $20k$ steps resulted in the HQI of $5.9$ ($\uparrow74.6\%$). The error in probe motion ($p$, $o$ and $f_z$) reduced by $31.6\%$, $26.6\%$ and $59.0\%$, with a mean error reduction of $\sim40\%$. The performance improved further when coaching was used more frequently, verifying its effectiveness for training of RUS.

% significantly high probe pose and force error of $12.3\%$ and $15.0\%$, respectively. Notably, only five coaching iterations (i.e, after every $20k$ step) reduced the probe pose and force error by $61.8\%$ and $76.7\%$, respectively. The HQI also increased from 2.1 to 8.2 ($\uparrow75.3\%$), which shows the effectiveness of coaching. 
% \vspace{-2mm}
\section{Conclusion and Future work}
This paper presents a coaching framework for RUS to improve the learning efficiency and accuracy of ultrasound image acquisition. {\color{black}Unlike previously proposed LfD methods for RUS, this framework leveraged real-time feedback from experts during the training process. Coaching is modeled as a Partially Observable Markov Decision Process (POMDP), which approximates the trajectory-based corrections from the expert to update the reward, objectives, and parameters of the DRL policy}. When tested for an ultrasound of urinary bladder phantom, this methodology improved the convergence rate of learned policy by $25\%$ and increased the number of high-quality image acquisitions by $74.4\%$. Future work will explore further improvement in performance through policy priors derived using offline expert demonstrations, as done in our previous works \cite{raina2023robotic, raina2023deep}.

\bibliography{references} 

\begin{thebibliography}{10}
\providecommand{\url}[1]{#1}
\csname url@samestyle\endcsname
\providecommand{\newblock}{\relax}
\providecommand{\bibinfo}[2]{#2}
\providecommand{\BIBentrySTDinterwordspacing}{\spaceskip=0pt\relax}
\providecommand{\BIBentryALTinterwordstretchfactor}{4}
\providecommand{\BIBentryALTinterwordspacing}{\spaceskip=\fontdimen2\font plus
\BIBentryALTinterwordstretchfactor\fontdimen3\font minus \fontdimen4\font\relax}
\providecommand{\BIBforeignlanguage}[2]{{%
\expandafter\ifx\csname l@#1\endcsname\relax
\typeout{** WARNING: IEEEtran.bst: No hyphenation pattern has been}%
\typeout{** loaded for the language `#1'. Using the pattern for}%
\typeout{** the default language instead.}%
\else
\language=\csname l@#1\endcsname
\fi
#2}}
\providecommand{\BIBdecl}{\relax}
\BIBdecl

\bibitem{carr2021influence}
J.~C. Carr, G.~R. Gerstner, C.~C. Voskuil, J.~E. Harden, D.~Dunnick, K.~M. Badillo, J.~I. Pagan, K.~K. Harmon, R.~M. Girts, J.~P. Beausejour \emph{et~al.}, ``The influence of sonographer experience on skeletal muscle image acquisition and analysis,'' \emph{Journal of functional morphology and kinesiology}, vol.~6, no.~4, p.~91, 2021.

\bibitem{raina2021comprehensive}
D.~Raina, H.~Singh, S.~K. Saha, C.~Arora, A.~Agarwal, S.~Chandrashekhara, K.~Rangarajan, and S.~Nandi, ``Comprehensive telerobotic ultrasound system for abdominal imaging: Development and in-vivo feasibility study,'' in \emph{2021 International Symposium on Medical Robotics (ISMR)}.\hskip 1em plus 0.5em minus 0.4em\relax IEEE, 2021, pp. 1--7.

\bibitem{chandrashekhara2022robotic}
S.~H. Chandrashekhara, K.~Rangarajan, A.~Agrawal, S.~Thulkar, S.~Gamanagatti, D.~Raina, S.~K. Saha, and C.~Arora, ``Robotic ultrasound: An initial feasibility study,'' \emph{World Journal of Methodology}, vol.~12, no.~4, p. 274, 2022.

\bibitem{jiang2023robotic}
Z.~Jiang, S.~E. Salcudean, and N.~Navab, ``Robotic ultrasound imaging: State-of-the-art and future perspectives,'' \emph{Medical image analysis}, p. 102878, 2023.

\bibitem{balakuntala2019extending}
M.~V. Balakuntala, V.~L. Venkatesh, J.~P. Bindu, R.~M. Voyles, and J.~Wachs, ``Extending policy from one-shot learning through coaching,'' in \emph{2019 28th IEEE International Conference on Robot and Human Interactive Communication (RO-MAN)}.\hskip 1em plus 0.5em minus 0.4em\relax IEEE, 2019, pp. 1--7.

\bibitem{haywood2021life}
K.~M. Haywood and N.~Getchell, \emph{Life span motor development}.\hskip 1em plus 0.5em minus 0.4em\relax Human kinetics, 2021.

\bibitem{pamies2023autonomous}
M.~B.~I. Pamies, M.~T. Villasevil, Z.~Wang, S.~Desai, P.~Agrawal, and A.~Gupta, ``Autonomous robotic reinforcement learning with asynchronous human feedback,'' in \emph{7th Annual Conference on Robot Learning}, 2023.

\bibitem{li2021autonomous}
K.~Li, J.~Wang, Y.~Xu, H.~Qin, D.~Liu, L.~Liu, and M.~Q.-H. Meng, ``Autonomous navigation of an ultrasound probe towards standard scan planes with deep reinforcement learning,'' in \emph{2021 IEEE International Conference on Robotics and Automation (ICRA)}.\hskip 1em plus 0.5em minus 0.4em\relax IEEE, 2021, pp. 8302--8308.

\bibitem{ning2021autonomic}
G.~Ning, X.~Zhang, and H.~Liao, ``Autonomic robotic ultrasound imaging system based on reinforcement learning,'' \emph{IEEE Transactions on Biomedical Engineering}, vol.~68, no.~9, pp. 2787--2797, 2021.

\bibitem{raina2023expert}
D.~Raina, D.~Ntentia, S.~Chandrashekhara, R.~Voyles, and S.~K. Saha, ``Expert-agnostic ultrasound image quality assessment using deep variational clustering,'' in \emph{2023 IEEE International Conference on Robotics and Automation (ICRA)}.\hskip 1em plus 0.5em minus 0.4em\relax IEEE, 2023, pp. 2717--2723.

\bibitem{jiang2023intelligent}
Z.~Jiang, Y.~Bi, M.~Zhou, Y.~Hu, M.~Burke, and N.~Navab, ``Intelligent robotic sonographer: Mutual information-based disentangled reward learning from few demonstrations,'' \emph{The International Journal of Robotics Research}, p. 02783649231223547, 2023.

\bibitem{mebarki20102}
R.~Mebarki, A.~Krupa, and F.~Chaumette, ``2-d ultrasound probe complete guidance by visual servoing using image moments,'' \emph{IEEE Transactions on Robotics}, vol.~26, no.~2, pp. 296--306, 2010.

\bibitem{nadeau2016moments}
C.~Nadeau, A.~Krupa, J.~Petr, and C.~Barillot, ``Moments-based ultrasound visual servoing: From a mono-to multiplane approach,'' \emph{IEEE Transactions on Robotics}, vol.~32, no.~6, pp. 1558--1564, 2016.

\bibitem{hennersperger2016towards}
C.~Hennersperger, B.~Fuerst, S.~Virga, O.~Zettinig, B.~Frisch, T.~Neff, and N.~Navab, ``Towards mri-based autonomous robotic us acquisitions: a first feasibility study,'' \emph{IEEE transactions on medical imaging}, vol.~36, no.~2, pp. 538--548, 2016.

\bibitem{al2021autonomous}
L.~Al-Zogbi, V.~Singh, B.~Teixeira, A.~Ahuja, P.~S. Bagherzadeh, A.~Kapoor, H.~Saeidi, T.~Fleiter, and A.~Krieger, ``Autonomous robotic point-of-care ultrasound imaging for monitoring of covid-19--induced pulmonary diseases,'' \emph{Frontiers in Robotics and AI}, vol.~8, p. 645756, 2021.

\bibitem{mylonas2013autonomous}
G.~P. Mylonas, P.~Giataganas, M.~Chaudery, V.~Vitiello, A.~Darzi, and G.-Z. Yang, ``Autonomous efast ultrasound scanning by a robotic manipulator using learning from demonstrations,'' in \emph{2013 IEEE/RSJ International Conference on Intelligent Robots and Systems}.\hskip 1em plus 0.5em minus 0.4em\relax IEEE, 2013, pp. 3251--3256.

\bibitem{li2022learning}
M.~Li and X.~Deng, ``Learning robotic ultrasound skills from human demonstrations,'' in \emph{Cognitive Robotics}.\hskip 1em plus 0.5em minus 0.4em\relax IntechOpen, 2022.

\bibitem{raina2023robotic}
D.~Raina, S.~Chandrashekhara, R.~Voyles, J.~Wachs, and S.~K. Saha, ``Robotic sonographer: Autonomous robotic ultrasound using domain expertise in bayesian optimization,'' in \emph{2023 IEEE International Conference on Robotics and Automation (ICRA)}.\hskip 1em plus 0.5em minus 0.4em\relax IEEE, 2023, pp. 6909--6915.

\bibitem{raina2023deep}
------, ``Deep kernel and image quality estimators for optimizing robotic ultrasound controller using bayesian optimization,'' in \emph{2023 International Symposium on Medical Robotics (ISMR)}.\hskip 1em plus 0.5em minus 0.4em\relax IEEE, 2023, pp. 1--7.

\bibitem{raina2023rusopt}
D.~Raina, A.~Mathur, R.~M. Voyles, J.~Wachs, S.~Chandrashekhara, and S.~K. Saha, ``Rusopt: Robotic ultrasound probe normalization with bayesian optimization for in-plane and out-plane scanning,'' in \emph{2023 IEEE 19th International Conference on Automation Science and Engineering (CASE)}.\hskip 1em plus 0.5em minus 0.4em\relax IEEE, 2023, pp. 1--7.

\bibitem{burke2023learning}
M.~Burke, K.~Lu, D.~Angelov, A.~Strai{\v{z}}ys, C.~Innes, K.~Subr, and S.~Ramamoorthy, ``Learning rewards from exploratory demonstrations using probabilistic temporal ranking,'' \emph{Autonomous Robots}, vol.~47, no.~6, pp. 733--751, 2023.

\bibitem{thomaz2005real}
A.~L. Thomaz, G.~Hoffman, and C.~Breazeal, ``Real-time interactive reinforcement learning for robots,'' in \emph{AAAI 2005 workshop on human comprehensible machine learning}, vol.~3, no. 3.7, 2005, p.~1.

\bibitem{macglashan2017interactive}
J.~MacGlashan, M.~K. Ho, R.~Loftin, B.~Peng, G.~Wang, D.~L. Roberts, M.~E. Taylor, and M.~L. Littman, ``Interactive learning from policy-dependent human feedback,'' in \emph{International conference on machine learning}.\hskip 1em plus 0.5em minus 0.4em\relax PMLR, 2017, pp. 2285--2294.

\bibitem{sadigh2017active}
D.~Sadigh, A.~D. Dragan, S.~Sastry, and S.~A. Seshia, \emph{Active preference-based learning of reward functions}, 2017.

\bibitem{biyik2018batch}
E.~Biyik and D.~Sadigh, ``Batch active preference-based learning of reward functions,'' in \emph{Conference on robot learning}.\hskip 1em plus 0.5em minus 0.4em\relax PMLR, 2018, pp. 519--528.

\bibitem{griffith2013policy}
S.~Griffith, K.~Subramanian, J.~Scholz, C.~L. Isbell, and A.~L. Thomaz, ``Policy shaping: Integrating human feedback with reinforcement learning,'' \emph{Advances in neural information processing systems}, vol.~26, 2013.

\bibitem{losey2019learning}
D.~P. Losey and M.~K. O'Malley, ``Learning the correct robot trajectory in real-time from physical human interactions,'' \emph{ACM Transactions on Human-Robot Interaction (THRI)}, vol.~9, no.~1, pp. 1--19, 2019.

\bibitem{bobu2020quantifying}
A.~Bobu, A.~Bajcsy, J.~F. Fisac, S.~Deglurkar, and A.~D. Dragan, ``Quantifying hypothesis space misspecification in learning from human--robot demonstrations and physical corrections,'' \emph{IEEE Transactions on Robotics}, vol.~36, no.~3, pp. 835--854, 2020.

\bibitem{li2021learning}
M.~Li, A.~Canberk, D.~P. Losey, and D.~Sadigh, ``Learning human objectives from sequences of physical corrections,'' in \emph{2021 IEEE International Conference on Robotics and Automation (ICRA)}.\hskip 1em plus 0.5em minus 0.4em\relax IEEE, 2021, pp. 2877--2883.

\bibitem{javdani2018shared}
S.~Javdani, H.~Admoni, S.~Pellegrinelli, S.~S. Srinivasa, and J.~A. Bagnell, ``Shared autonomy via hindsight optimization for teleoperation and teaming,'' \emph{The International Journal of Robotics Research}, vol.~37, no.~7, pp. 717--742, 2018.

\bibitem{haarnoja2018soft}
T.~Haarnoja, A.~Zhou, P.~Abbeel, and S.~Levine, ``Soft actor-critic: Off-policy maximum entropy deep reinforcement learning with a stochastic actor,'' in \emph{International conference on machine learning}.\hskip 1em plus 0.5em minus 0.4em\relax PMLR, 2018, pp. 1861--1870.

\bibitem{singh2019end}
A.~Singh, L.~Yang, C.~Finn, and S.~Levine, ``End-to-end robotic reinforcement learning without reward engineering,'' \emph{Robotics: Science and Systems XV}, 2019.

\bibitem{raina2024deep}
D.~Raina, S.~Chandrashekhara, R.~Voyles, J.~Wachs, and S.~K. Saha, ``Deep learning model for quality assessment of urinary bladder ultrasound images using multi-scale and higher-order processing,'' \emph{IEEE Transactions on Ultrasonics, Ferroelectrics, and Frequency Control}, 2024.

\bibitem{chen2016pomdp}
M.~Chen, E.~Frazzoli, D.~Hsu, and W.~S. Lee, ``Pomdp-lite for robust robot planning under uncertainty,'' in \emph{2016 IEEE International Conference on Robotics and Automation (ICRA)}.\hskip 1em plus 0.5em minus 0.4em\relax IEEE, 2016, pp. 5427--5433.

\bibitem{losey2022physical}
D.~P. Losey, A.~Bajcsy, M.~K. O’Malley, and A.~D. Dragan, ``Physical interaction as communication: Learning robot objectives online from human corrections,'' \emph{The International Journal of Robotics Research}, vol.~41, no.~1, pp. 20--44, 2022.

\end{thebibliography}
\bibliographystyle{IEEEtran}

\end{document}